\documentclass[11pt]{article}
\usepackage{coling2020}
\usepackage{times}
\usepackage{url}
\usepackage{latexsym}

\usepackage{enumitem}
\usepackage{microtype}

\usepackage{graphicx} 
\usepackage{multirow}
\usepackage{tabularx}
\usepackage{booktabs}
\usepackage{latexsym}
\usepackage{amsmath}
\usepackage{amssymb}
\usepackage{mathrsfs}
\usepackage{enumerate}
\usepackage{bm}
\usepackage{paralist}
\usepackage{subfigure}
\usepackage{caption}
\usepackage{fancyhdr}
\usepackage{courier}
\usepackage{color,xcolor,colortbl}
\usepackage{array}
\usepackage{color}
\usepackage[ruled,vlined]{algorithm2e}
\newcommand{\textcite}[1]{\citeauthor{#1} (\citeyear{#1})}
\newcommand{\func}[1]{\mathit{#1}}

\title{When does MAML Work the Best? An Empirical Study on Model-Agnostic Meta-Learning in NLP Applications}

\author{Zequn Liu, Ruiyi Zhang, Yiping Song, Wei Ju, Ming Zhang\thanks{$^*$Corresponding author} \\
  Department of Computer Science, School of EECS, Peking University, Beijing, China \\
  \texttt{{zequnliu,zhangruiyi,songyiping,juwei,mzhang\_cs}@pku.edu.cn}\\}

\date{}

\begin{document}
\maketitle
\begin{abstract}
Model-Agnostic Meta-Learning (MAML) is successfully employed in Natural Language Processing applications including few-shot text classification and multi-domain low-resource language generation. Many impacting factors, including data quantity and data distribution, can affect the performance of MAML in NLP, but few works have thoroughly studied them. In this paper, we conduct an empirical study to investigate these impacting factors and conclude when MAML works the best based on the experimental results.
\end{abstract}

\section{Introduction}
In the field of Natural Language Processing (NLP), the abundance of training data plays a crucial role in the performance of deep learning models \cite{dodge2021documenting}. However, numerous NLP applications face a substantial challenge due to the scarcity of annotated data \cite{schick2021few}. For example, in personalized dialogue generation, each user's annotated data is limited, making it difficult to train a well-performing response generation model for each individual \cite{paml,yiping}. 
Many techniques have been employed to address the issue of data scarcity, including self-supervised pre-training \cite{achiam2023gpt,chatGPT}, transfer learning \cite{gero2018transfer,kumar2022unsupervised}, and meta-learning \cite{paml,yiping,zhao2022improving}. Compared to other approaches, meta-learning focuses on designing models that learn to learn from small data sets, reducing the dependency on extensive pre-training data \cite{maml,vilalta2002perspective}. Therefore, meta-learning has been widely applied in low-resource NLP tasks. 

Model-Agnostic Meta-Learning (MAML)~\cite{maml} is one of the most popular meta-learning methods. It is trained on plenty of tasks (i.e. small data sets) to get a parameter initialization which is easy to adapt to target tasks with a few samples. As a model-agnostic framework, MAML is successfully employed in different NLP applications.
Some works use MAML for few-shot text classification, such as relation classification~\cite{obamuyide2019model} and topic classification~\cite{bao2019few}.
Other works use MAML for multi-domain and low-resource language generation, such as few-shot dialogue system~\cite{mi2019meta,paml,yuzhou,yiping} and low-resource machine translation~\cite{translation}.

When applying MAML to NLP, several factors can influence the training strategy and performance of the model. Firstly, the \textbf{data quantity} within the datasets used as "tasks" varies across different applications, which can impact the effectiveness of MAML \cite{dataset,yiping}. Secondly, while vanilla MAML assumes that the \textbf{data distribution} is the same across tasks, in real-world NLP tasks, the data distributions can differ significantly \cite{li2018learning,balaji2018metareg}. For example, PAML \cite{paml} regards each person's dialogues as a task for MAML and they have different personal profiles. This variation manifests both between training tasks and between training and testing tasks, similarly affecting the performance of MAML. Few works have thoroughly studied these impact factors.

In this paper, we take an empirical approach to systematically investigating these impacting factors and finding when MAML works the best. We conduct extensive experiments over 4 datasets. We first study the effects of data quantity and distribution on the training strategy:
\textit{\textbf{RQ1}. Since the parameter initialization learned by MAML can be seen as a general language model of training tasks, when the training and testing tasks have different data distributions, how can the general language model training affect the model's task-specific adaptation ability?
\textbf{RQ2}. How do the data distribution and data quantity affect our decision of fine-tuning epoch number?}
Then we study the effects of these factors on the model performance:
\textit{\textbf{RQ3}. How do the data quantity and data distribution affect the performance of MAML?}


The experimental results provide insights on MAML in NLP applications. Our conclusions help researchers to better develop meta-learning methods in NLP.

\section{Preliminaries}
\subsection{Meta-learning Problem Definition}
In meta-learning, we have multiple tasks $T$ sampled from distribution $p(\mathcal{T})$~\cite{ravi2016optimization,andrychowicz2016learning,santoro2016meta}. For each task $T_i$, we train a \textbf {base model} $f_i^\theta$ with parameter $\theta_i$ on its training corpus $D_i^{train}$ which only has a few samples, and evaluate the model on the testing corpus $D_i^{valid}$. We divide the tasks into \textit{meta-training}, \textit{meta-validation}, and \textit{meta-testing}. The goal of meta-learning is that after training on \textit{meta-training}, we can quickly find $f_i^\theta$ via \textbf{fine-tuning (adaptation)} with $D_i^{train}$ for each task $T_i$ in \textit{meta-testing}.
\subsection{Model-Agnostic Meta-Learning}
Model-Agnostic Meta-Learning (MAML) \cite{maml} pre-trains a parameter initialization $\theta$ shared among tasks.
At each training epoch, MAML samples a set of tasks ${T_i}{\sim} p(\mathcal{T})$. For each task $T_i$, MAML trains from the initialization $\theta$ on $D_i^{train}$ to get $\theta_i$, then evaluates each $\theta_i$ on $D_i^{valid}$ and updates $\theta$, which is,
\begin{equation}
\begin{aligned}
\label{euq: meta-training}
\theta_i = \theta - \alpha \nabla_{\theta}\mathcal{L}_{D_i^{train}}(\theta), 
\theta = \theta - \beta \nabla_{\theta}\sum_{{T_i}{\sim} p(\mathcal{T})}\mathcal{L}_{D_i^{valid}}(\theta_i)
\end{aligned}
\end{equation}
where $\mathcal{L}_{D_i^{train}}(\theta)$ and $\mathcal{L}_{D_i^{valid}}(\theta_i)$ are the loss functions of $\theta$ on $D_i^{train}$ and $\theta_i$ on $D_i^{valid}$,  $\alpha$ and $\beta$ are the learning rates. 
In fine-tuning stage, each task fine-tunes from the pre-trained initialization $\theta$ on its $D_i^{train}$.
\section{Experimental Setup}
\subsection{Datasets}
In \textbf{Experiment I: Text Classification}, we use {\tt FewRel}~\cite{han2018fewrel} and {\tt Amazon}~\cite{he2016ups}. They are datasets for 5-way 5-shot classification, which means 5 classes are randomly sampled from the full dataset for each task, and each class has 5 samples. {\tt FewRel} is a relation classification dataset with 65/5/10 tasks for meta-training/meta-validation/meta-testing. 
{\tt Amazon} is a large customer review dataset with 24 product categories. We follow \cite{bao2019few} to sample 1000 reviews from each category and use 10/5/9 tasks for meta-training/meta-validation/meta-testing.

In \textbf{Experiment II: Dialogue Generation}, we use {\tt Persona}~\cite{persona} and {\tt Weibo}, regarding building a dialogue model for a user as a task. {\tt Persona} is a personalized dialogue dataset with 1137/99/100 users for meta-training/meta-validation/meta-testing. Each user has 121 utterances on average. {\tt Weibo} is a personalized dialogue dataset collected from Weibo conversations with 371/40/38 users for meta-training/meta-validation/meta-testing. Each user has 1200 utterances on average.





\subsection{Base Models and Settings}
In each experiment, we use the same base model among the 2 datasets.
In text classification experiments, we use the CNN proposed in \cite{bao2019few} as the base model and follow the hyperparameter settings.
 We use Transformer~\cite{vaswani2017attention} as the base model in dialogue generation experiment. 
 In {\tt Persona}, we use pre-trained Glove embedding~\cite{glove}. In {\tt Weibo}, we use Gensim~\cite{gensim}. We follow the other hyperparameter settings in \cite{paml}.
\subsection{Evaluation Metrics}
~\label{sec:eval}
In text classification experiment, we use accuracy (Acc) to evaluate the classification performance.
In dialogue generation experiment, we evaluate the performance of MAML in terms of quality and personality. We use PPL and BLEU~\cite{bleu} to measure the similarity between the reference and the generated response, and use C Score~\cite{paml} to measure the personality. In {\tt Persona} we use a pre-trained natural language inference model to measure the response consistency with persona description for C Score. In {\tt Weibo}, users do not have persona descriptions, so we pre-train a user classifier to classify the generated responses, and use the accuracy for C Score.

\section{Results and Analysis}
\subsection{Trade-off Problem between General Language Model and Task-specific Adaptation}
\begin{figure*}[!t]
\centering
\setlength{\belowcaptionskip}{0pt}
\setlength{\abovecaptionskip}{0pt}
\subfigure[\vspace{0pt}C Score and BLEU on \tt{Persona}\vspace{0pt}]{
\begin{minipage}[b]{0.3\linewidth}
\centering
\includegraphics[width=\textwidth]{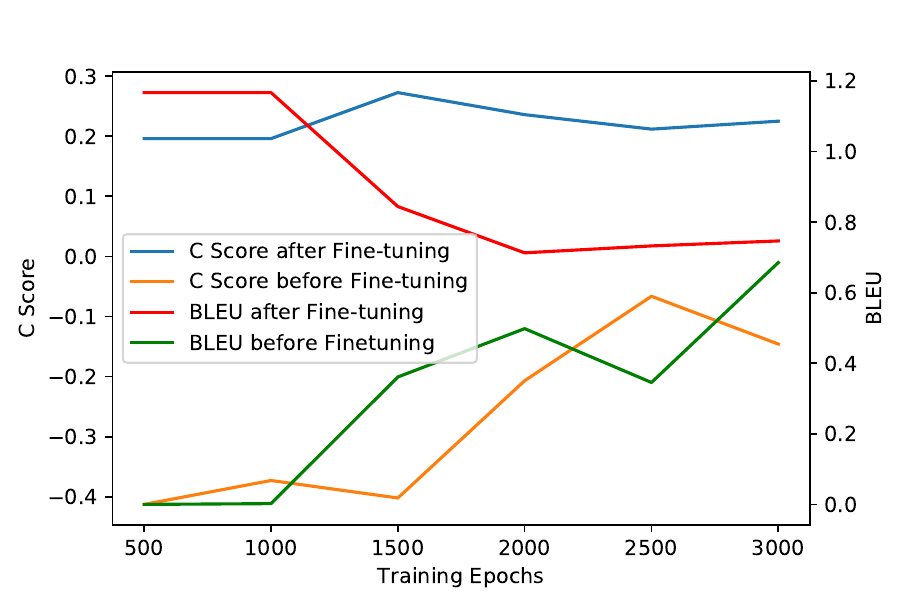}
\label{persona_c}
\end{minipage}
}
\subfigure[\vspace{0pt}C Score and BLEU on \tt{Weibo}\vspace{0pt}]{
\begin{minipage}[b]{0.3\linewidth}
\centering
\includegraphics[width=\textwidth]{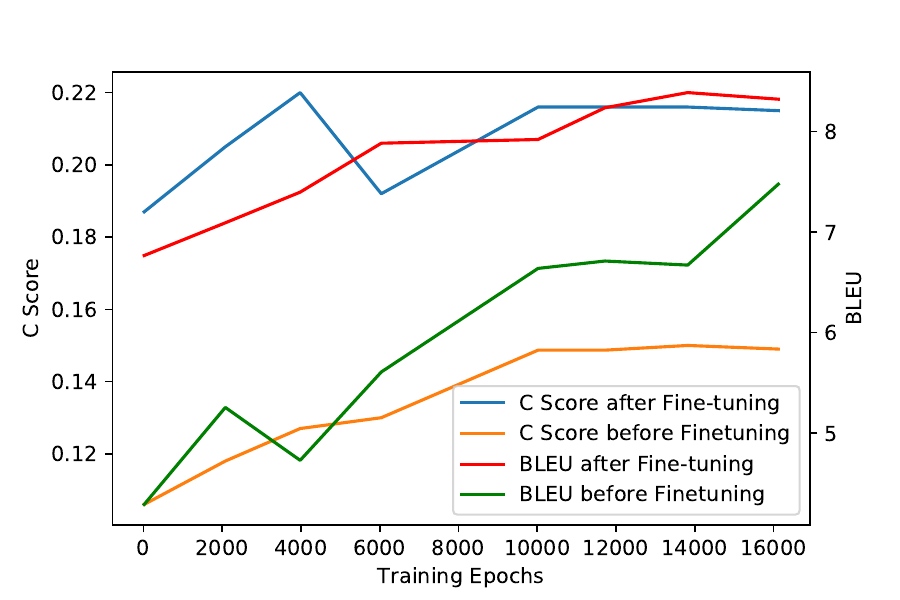}
\label{weibo_c}
\end{minipage}
}
\subfigure[\vspace{0pt}Acc on \tt{Amazon} and \tt{FewRel}\vspace{0pt}]{
\begin{minipage}[b]{0.3\linewidth}
\centering
\includegraphics[width=\textwidth]{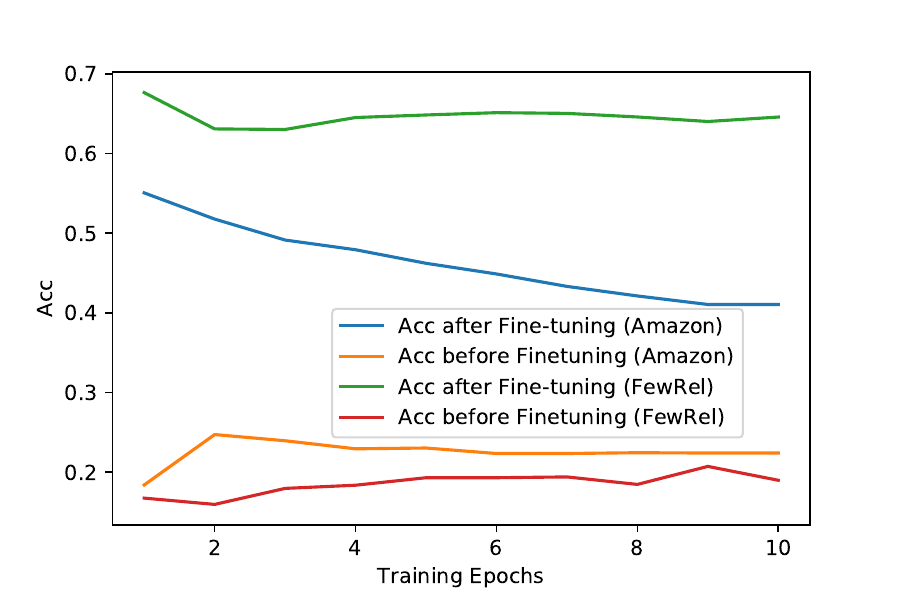}
\label{amazon}
\end{minipage}
}
\caption{The changing trend of the performance (accuracy for text classification, personality (C Score) and quality (BLEU) for dialogue genration) before and after fine-tuning as the training epochs increase.}
\label{fig:trend}
\end{figure*}

To answer RQ1, we compare the changing trend of the general language model and the task-specific adaptation ability during the training of MAML to find whether there is a trade-off problem. (Figure \ref{fig:trend}) We select the trained parameter initialization at different MAML training epochs and evaluate them directly on the meta-testing set before fine-tuning, using the quality performance (accuracy for classification and BLEU for generation) to 
evaluate the general language model. Then we use the personality performance (accuracy for classification and C Score for generation) after fine-tuning to measure each model's ability to adapt to specific tasks. 

When the training epochs increase, the model's ability of task-specific adaptation reaches the peak earlier than the quality of its general language model does. 
In {\tt Persona},
BLEU before fine-tuning reaches the peak at epoch 3000,
but after epoch 1500, the final C Score drops.
In {\tt Weibo}, the changing trends are consistent with the results on {\tt Persona}.
In {\tt FewRel} and {\tt Amazon}, the general language model first becomes better and then over-fits to the training data, but the final accuracy decreases after epoch 1.

The finding suggests that parameter initialization at the late training stage has strong general language generation ability, but performs comparative poorly in task-specific adaptation. 
Although in the early training stage, the performance improves benefiting from the pre-trained general language model, if the language model becomes too ``general", it will lose the ability of adapting to specific tasks. It is noteworthy that the "too general" problem is not the same as over-fitting, since the "too general" model performs well before fine-tuning, which means it does not over-fit to the training data.
\subsection{Impact of Data Quantity and Task Profile on Fine-tuning}
To answer RQ2, we find the fine-tuning epochs for each task in {\tt Persona} where its BLEU and C Score reaches the best respectively to find the impact of data quantity and the task profile (persona description) on fine-tuning. (Table~\ref{tab:fine-tune}) We cluster the tasks with similar best fine-tuning epoch number and calculate the average dialogue quantity and task profile similarity for each cluster.  We define Jac Score of the persona descriptions to evaluate their task profile similarity as $\func{Jac~Score = \Sigma_{i=1}^N Jac_i / (NJac_{whole})}$,
where $Jac_i$ is the Jaccard similarity of the persona descriptions in cluster i, $Jac_{whole}$ is the Jaccard similarity of all the persona descriptions, and N is the number of clusters.
If Jac Score is larger than 1, the persona descriptions are similar to each other in each cluster. Larger Jac Score means better similarity.

For both BLEU and C Score, Jac Score is around 1 in each cluster, which means the persona descriptions are not similar. The dialogue quantity also seems similar among different clusters. So we can conclude that data quantity and task profile does not have a major impact on the fine-tuning process.

\begin{table*}[!t]
\small
\centering
\setlength{\belowcaptionskip}{-10pt}
\setlength{\abovecaptionskip}{0pt}
\begin{tabular}{lp{20mm}p{13mm}p{13mm}|p{20mm}p{13mm}p{13mm}}
\hline
\textbf{\multirow{2}{*}{Cluster}} & \multicolumn{3}{c|}{\textbf{BLEU}} & \multicolumn{3}{c}{\textbf{C Score}} \\
\cline{2-4}\cline{5-7}
& \textbf{Mean Fine-tuning Epochs} & \textbf{Task Similarity} & \textbf{Dialogue Quantity}  & \textbf{Mean Fine-tuning Epochs} & \textbf{Task Similarity} & \textbf{Dialogue Quantity}\\
\hline
1&0.22&1.05&9.31&2.50&1.00&9.05\\
2&1.16&1.04&9.63&5.83&1.06&9.89\\
3&2.07&1.01&9.94&8.86&1.09&10.05\\
4&3.68&1.00&9.63&10.25&1.05&9.16\\
5&5.58&1.00&9.85&12.86&1.00&10.24\\
\hline
\end{tabular}

\caption{The average dialogue quantity and task profile similarity of the task clusters grouped by each task's  ``best fine-tuning epochs" of BLEU and C (where its BLEU and C reaches the best) in {\tt Persona}. }
\label{tab:fine-tune}
\end{table*}

\subsection{Impact of Data Quantity and Task Similarity on the performance of MAML}
To answer RQ3, we conduct experiments on different data quantity and task similarity settings. We compare two baselines with \textbf{MAML} :
\textbf{Transformer/CNN}, which pre-trains the base model (Transformer/CNN) on the meta-training set and evaluates directly on the meta-testing set, and \textbf{Transformer/CNN-F}, which fine-tunes \textbf{Transformer/CNN} on each meta-testing task.

\textbf{Data Quantity.} In {\tt Persona}, we evaluate \textbf{Transformer/CNN}, \textbf{Transformer/CNN-F} and \textbf{MAML} on 3 data quantity settings: 50/100/120-shot (each task has 50, 100, 120 utterances on average).  In {\tt Weibo}, {\tt FewRel} and {\tt Amazon}, the settings are 500/1000/1500-shot, 3/4/5-shot and 3/4/5-shot respectively (Table \ref{tab:overall}).
When the data quantity is small, the advantage of MAML is more significant. In {\tt Persona}, the C Score and BLEU of \textbf{MAML} outperform baselines on 50-shot and 100-shot settings, but on 120-shot setting, the BLEU of \textbf{MAML} is lower than \textbf{Transformer-F}. In {\tt Weibo}, {\tt FewRel} and {\tt Amazon}, the percentages that \textbf{MAML} outperforms the baselines by also decrease as the data quantity increasing. This finding is in line with the mechanism of MAML. MAML finds a sensitive parameter initialization that can adapt with few data samples~\cite{maml}.
When there are enough data samples, fine-tuning also performs well, so BLEU of \textbf{Transformer-F} in {\tt Persona} on 120-shot setting is even better.

\begin{table*}[!t]
\small
\centering
\setlength{\belowcaptionskip}{-10pt}
\setlength{\abovecaptionskip}{0pt}
\begin{tabular}{lccc|ccc|c|c}
\hline
\textbf{\multirow{2}{*}{Method}} & \multicolumn{3}{c|}{\textbf{Persona}} & \multicolumn{3}{c|}{\textbf{Weibo}} &
\multicolumn{1}{c|}{\textbf{FewRel}} & \multicolumn{1}{c}{\textbf{Amazon}} \\
\cline{2-4}\cline{5-7}\cline{8-9}
& \textbf{PPL} & \textbf{C Score} & \textbf{BLEU}  & \textbf{PPL} & \textbf{C Score} & \textbf{BLEU}  & \textbf{Acc} & \textbf{Acc}\\
\hline
&\textbf{50-shot} &&&\textbf{500-shot}&&&\textbf{3-shot}&\textbf{3-shot}\\
\hline
Transformer/CNN &96.56	&-0.07 &0.11&36.92	&0.09	&1.94&0.17&0.18\\
Transformer/CNN-F  &\textbf{48.46}	&-0.04	&0.46&\textbf{34.21}	&0.15	&2.50&0.54&0.50\\
MAML &55.45	&\textbf{0.10}	&\textbf{0.55}&53.01	&\textbf{0.21}	&\textbf{6.09}&\textbf{0.61}&\textbf{0.51}\\
\hline
&\textbf{100-shot} &&&\textbf{1000-shot}&&&\textbf{4-shot}&\textbf{4-shot}\\
\hline
Transformer/CNN &36.83	&-0.08	&0.75&61.81	&0.10&	2.92&0.18&0.21\\
Transformer/CNN-F  &\textbf{31.21}	&-0.01	&0.68&52.88	&0.17	&3.42&0.59&0.54\\
MAML &43.57	&\textbf{0.19}	&\textbf{0.82}&\textbf{47.62}	&\textbf{0.22}	&\textbf{6.75}&\textbf{0.65}&\textbf{0.55}\\
\hline
&\textbf{120-shot} &&&\textbf{1200-shot}&&&\textbf{5-shot}&\textbf{5-shot}\\
\hline
Transformer/CNN &36.54	&-0.03	&0.80&36.95&0.10&2.87&0.20&0.20\\
Transformer/CNN-F  &\textbf{34.28}	&0.08	&0.89&\textbf{35.02}&0.20&5.17&0.66&\textbf{0.55}\\
MAML &56.21	&\textbf{0.27}	&\textbf{0.84}&41.88&\textbf{0.22}&\textbf{8.39}&\textbf{0.68}&\textbf{0.55}\\
\hline
&\textbf{Similar} &&&\textbf{Similar}&&&-&-\\
\hline
Transformer/CNN-F  &\textbf{35.93}	&-0.03	&\textbf{0.70}&\textbf{90.93}	&\textbf{0.05}	&0.33&-&-\\
MAML &42.50	&\textbf{-0.02}&0.63&92.79	&0.04	&\textbf{0.38}	&-&-\\
\hline
\end{tabular}

\caption{The performance on different data quantity and task similarity settings.}
\label{tab:overall}
\end{table*}

\textbf{Task similarity.} In {\tt Persona} and {\tt Weibo}, each task is a set of dialogues for one user, so tasks are different from each other. We shuffle the samples and randomly divide tasks to construct the setting that tasks are similar to each other. For a fair comparison, each task on this setting also has 120 and 1200 utterances on average in {\tt Persona} and {\tt Weibo} respectively. We train and evaluate \textbf{Transformer-F} and \textbf{MAML} on this setting. (Table \ref{tab:overall}).
When tasks are similar to each other, \textbf{MAML} performs comparatively poorly. In {\tt Persona} and {\tt Weibo}, the performance of \textbf{MAML} is similar to that of \textbf{Transformer-F}, while \textbf{MAML} performs significantly better than \textbf{Transformer-F} when tasks are different. A possible explanation is that if there is no clear distinction between tasks, the meta-learning setting can be viewed as a transfer learning setting, which only has a source domain and a target domain, and fine-tuning performs well in transfer learning. 
So if the tasks are similar to each other, we can simply use \textbf{Transformer-F} rather than \textbf{MAML}.

\section{Conclusion}
In this paper, we conduct an empirical study to investigate the impacting factors on the performance of MAML in NLP applications. We show that MAML works the best when the general language model is not fully trained by MAML, the data quantity of each task is small and tasks are dissimilar with each other. We also point out that it is unnecessary to customize the fine-tuning epoch number for each task according to the task profile or data quantity. Our work sheds light on the application of  MAML in NLP.
\bibliographystyle{acl}
\bibliography{coling2020}

\end{document}